\documentclass[runningheads,a4]{llncs}
\usepackage{amssymb}
\usepackage{float}
\usepackage{graphicx}
\usepackage{subfigure}
\usepackage{algorithm}
\usepackage{algorithmicx}
\usepackage[noend]{algpseudocode}
\usepackage{amsmath}
\usepackage{CJK}
\usepackage{verbatim}
\usepackage{url}
\usepackage{caption}
\begin{document}
\title{Assessing the Potential of Classical Q-learning in General Game Playing}
%
%
\author{Hui Wang, Michael Emmerich, Aske Plaat}
\authorrunning{Hui Wang et al.}
%
\institute{Leiden Institute of Advanced Computer Science, Leiden University,\\ Leiden, the Netherlands\\
\email{h.wang.13@liacs.leidenuniv.nl}\\
\url{http://www.cs.leiden.edu}}
\maketitle              

\begin{abstract}
  After the recent groundbreaking results of AlphaGo and AlphaZero, we have seen strong interests in deep reinforcement learning and artificial general intelligence (AGI) in game playing. However, deep learning is resource-intensive and the theory is not yet well developed. For small games, simple classical table-based Q-learning might still be the algorithm of choice. General Game Playing (GGP) provides a good testbed for reinforcement learning to research AGI. Q-learning is one of the canonical reinforcement learning methods, and has been used by (Banerjee $\&$ Stone, IJCAI 2007) in GGP.
  In this paper we implement Q-learning in GGP for three small-board games (Tic-Tac-Toe, Connect Four, Hex)\footnote{source code: https://github.com/wh1992v/ggp-rl}, to allow comparison to Banerjee et al.. We find that Q-learning converges to a high win rate in GGP. For the $\epsilon$-greedy strategy, we propose a first enhancement, the dynamic $\epsilon$ algorithm.
  In addition, inspired by (Gelly $\&$ Silver, ICML 2007) we combine online search (Monte Carlo Search) to enhance offline learning, and propose QM-learning for GGP. Both enhancements improve the performance of classical Q-learning.
  In this work, GGP allows us to show, if augmented by appropriate enhancements, that classical table-based Q-learning can perform well in small games. 

\end{abstract}

\section{Introduction}
Traditional game playing programs are written to play a single specific game, such as Chess, or Go. The aim of {\em General\/} Game Playing~\cite{Genesereth2005} (GGP) is to create adaptive game playing programs; programs that can play more than one game well. To this end, GGP uses a so-called Game Description Language (GDL)~\cite{Love2008}.
GDL-authors write game-descriptions that specify the rules of a game. The challenge for GGP-authors is to write a GGP player that will play  any game well.
GGP players should ensure that a wide range of GDL-games can be played well. Comprehensive tool-suites exist to help researchers write GGP and GDL programs, and an active research community exists~\cite{Kaiser2007,Genesereth2014,Swiechowski2014}.

The GGP model follows the state/action/result paradigm of \mbox{reinforcement} learning~\cite{Sutton1998}, a paradigm that has yielded many successful problem solving algorithms. For example, the successes of AlphaGo are based on two reinforcement learning algorithms, Monte Carlo Tree Search (MCTS)~\cite{Browne2012} and Deep Q-learning (DQN)~\cite{Mnih2015,Silver2016}. MCTS, in particular, has been successful in GGP~\cite{Mehat2008}. However, few works analyze the potential of Q-learning for GGP, not to mention DQN. The aim of this paper is to be a basis for further research of DQN for GGP.

Q-learning with deep neural networks requires extensive computational resources. Table-based Q-learning might offer a viable alternative for small games. Therefore, following Banerjee~\cite{Banerjee2007}, in this paper we address the convergence speed of table-based Q-learning. We use three small two-player zero-sum games: Tic-Tac-Toe, Hex and Connect Four, and table-based Q-learning. We introduce two enhancements: dynamic $\epsilon$, and, borrowing an idea from~\cite{Gelly2007}, we create a new version of Q-learning, 
inserting Monte Carlo Search (MCS) into Q-learning, using online search for offline learning.


Our contributions can be summarized as follows:
\begin{enumerate}
\item \textbf{Dynamic $\epsilon$:} We evaluate the classical Q-learning, finding (1) that Q-learning works and converges in GGP, and (2) that Q-learning with a dynamic $\epsilon$ can enhance the performance of TD($\lambda$) baseline with a fixed $\epsilon$~\cite{Banerjee2007}.
\item \textbf{QM-learning:} To further improve performance we enhance classical Q-learning by adding a modest amount of Monte Carlo lookahead (QMPlayer)~\cite{Robert2004}. This improves the convergence rate of Q-learning, and shows that online search can also improve the offline learning in GGP.
\end{enumerate}

The paper is organized as follows. Section 2 presents  related work and recalls basic concepts of GGP and reinforcement learning. 
Section 3 presents the designs of the QPlayer with fixed and dynamic $\epsilon$ and QMPlayer for two-player zero-sum games for GGP to assess the potential of classical Q-learning in detail.
Section 4 presents the experimental results. Section 5 concludes the paper and discusses directions for future work.

\section{Related Work and Preliminaries}
\subsection{GGP}
A General Game Player must be able to accept formal GDL descriptions of a game and play games effectively without human intervention~\cite{Genesereth2014}, where the GDL has been defined to describe the game rules~\cite{Thielscher2011}. An interpreter program~\cite{Swiechowski2014} generates the legal moves (actions) for a specific board (state).
Furthermore, a Game Manager (GM) is at the center of the software ecosystem. The GM interacts with game players through the TCP/IP protocol to control the  match. The GM manages game descriptions and matches records and temporary states of matches while the game is running. The system also contains  a viewer interface for users who are interested in running matches and a monitor to analyze the match process.

\subsection{Reinforcement Learning}
Since Watkins proposed Q-learning in 1989~\cite{Watkins1989}, much progress has been made in reinforcement learning~\cite{Even-Dar2002,Hu2003}. However,  few works report on the use of Q-learning in GGP. In~\cite{Banerjee2007}, Banerjee and  Stone propose a method to create a general game player to study knowledge transfer, combining Q-learning and GGP. Their aim is to improve the performance of Q-learning by transferring the knowledge learned in one game to a new, but related, game. They found knowledge transfer with Q-learning to be expensive. In~\cite{Gelly2007}, Gelly and Silver combine online and offline knowledge to improve learning performance. 

Recently, DeepMind published work on mastering Chess and Shogi by self-play with a deep, generalized reinforcement learning algorithm~\cite{Silver2017a}. With a series of landmark publications from AlphaGo to AlphaZero~\cite{Silver2016,Silver2017a,Silver2017b}, these \mbox{works} showcase the promise of general reinforcement learning algorithms. However, such learning algorithms are very resource-intensive and typically require special GPU/TPU hardware. Furthermore, the neural network-based approach is quite inaccessible to theoretical analysis. Therefore, in  this paper we study performance of table-based Q-learning.

In General Game Playing, variants of MCTS~\cite{Browne2012} are used with great success~\cite{Mehat2008}. M\'ehat et al. combined UCT and nested MCS for single-player general game playing~\cite{Mehat2010}. Cazenave et al. further proposed a nested MCS for two-player games~\cite{Cazenave2016}. Monte Carlo techniques have proved a viable approach for searching intractable game spaces and other optimization problems~\cite{Ruijl2014}. Therefore, in this paper we combine MCS to improve performance.

\subsection{Q-learning}
A basic distinction between reinforcement learning methods is that of  "on-policy" and "off-policy" methods. On-policy methods attempt to evaluate or improve the policy that is used to make decisions, whereas off-policy methods evaluate or improve a policy {\em different\/} from that used to make decisions~\cite{Sutton1998}. Q-learning is an off-policy method. The reinforcement learning model consists of an \emph{agent}, a set of states $S$, and a set of actions $A$ available in state $S$~\cite{Sutton1998}. The agent can move to the next state $s^\prime$, $s^\prime\in S$ from state $s$ after following action $a$, $a\in A$, denoted as $s\xrightarrow{a}s^\prime$. After finishing the action $a$, the agent gets an immediate reward $R(s, a)$, usually a numerical score. The cumulative return of current state $s$ by taking the action $a$, denoted as $Q(s, a)$, is a weighted sum, calculated by $R(s, a)$ and the maximum $Q(s^\prime,a^\prime)$ value of all next states:
\begin{equation}
Q(s,a)=R(s,a)+\gamma\ max_{a^\prime}Q(s^\prime,a^\prime)
\end{equation}
where $a^\prime \in A^\prime$ and $A^\prime$ is the set of actions available in state $s^\prime$. $\gamma$ is the discount factor of $max_{a^\prime}Q(s^\prime,a^\prime)$ for next state $s^\prime$. $Q(s,a)$ can be updated by online interactions with the environment using
the following rule:
\begin{equation}
Q(s,a)\leftarrow (1-\alpha)\ Q(s,a)+\alpha\ (\ R(s,a)+\gamma\ max_{a^\prime}Q(s^\prime,a^\prime))
\end{equation}
where $\alpha \in [0,1]$ is the learning rate. The Q-values are guaranteed to converge after iteratively updating.

\section{Design}
\subsection{Classical Q-learning for Two-Player Games}
GGP games in our experiments are two-player zero-sum games that alternate moves. Therefore, we can use the same rule, see Algorithm~\ref{alg:algorithmclassicalQ} line~\ref{alg:algorithmclassicalQ:rule}, to create $R(s,a)$, rather than to use a reward table. In our experiments, we set $R(s,a)=0$ for non-terminal states, and call the $getGoal()$ function for terminal states. 
In order to improve the learning effectiveness, we update the $Q(s,a)$ table only at the end of the match. During offline learning, QPlayer uses an $\epsilon$-greedy strategy to balance exploration and exploitation towards convergence. While the $\epsilon$-greedy strategy is enabled, QPlayer will perform a random action. Otherwise, QPlayer will perform the best action according to Q(S,A) table. If no record matches current state, QPlayer will perform a random action. The pseudo code for this algorithm is given in Algorithm~\ref{alg:algorithmclassicalQ}.
\allowdisplaybreaks
\begin{algorithm}[H]
\caption{Classical Q-learning Player with Static $\epsilon$}
\label{alg:algorithmclassicalQ}
\begin{algorithmic}[1] 
\Function{QPlayer}{current state $s$, learning rate $\alpha$, discount factor $\gamma$,
Q table: $Q(S,A)$}
\For{each match}
\If{$s$ terminates}
\For{each (s, a) from end to the start in current match record}
\State R(s,a)\ =$s^\prime$ is terminal state?\ getGoal($s^\prime$, myrole)\ :\ 0\label{alg:algorithmclassicalQ:rule}
\State Update $Q(s,a)\leftarrow (1-\alpha)\ Q(s,a)+\alpha\ (\ R(s,a)+\gamma\ max_{a^\prime}Q(s^\prime , a^\prime))$
\EndFor
\Else
\If{$\epsilon$-greedy is enabled}
\State selected\_action\ =\ Random()
\Else
\State selected\_action\ =\ SelectFromQTable()
\If{no $s$ record in $Q(S,A)$}
\State \emph{\textbf{selected\_action\ =\ Random()}}  \label{alg:algorithmclassicalQ:random}
\State \Comment{To be changed for different versions}
\EndIf
\EndIf
\State performAction($s$, selected\_action)
\EndIf
\EndFor
\State \Return $Q(S,A)$ 
\EndFunction
\end{algorithmic}
\end{algorithm}

\subsection{Dynamic $\epsilon$ Enhancement}
In contrast to the baseline of~\cite{Banerjee2007}, which uses a fixed $\epsilon$ value, we use a dynamically decreasing $\epsilon$-greedy Q-learning~\cite{Even-Dar2002}. In our implementation, we use the function
\begin{equation}\label{equation3}
\epsilon(m)=\begin{cases}
a(\cos(\frac{m}{2l}\pi))+b & m\leq l\\
0& m>l
\end{cases}
\end{equation}
for $\epsilon$, where $m$ is the current match count, and $l$ is a number of matches we set in advance to control the decaying speed of $\epsilon$. During offline learning, if $m=l$, $\epsilon$ decreases to 0. $a$ and $b$ is set to limit the range of $\epsilon$, where $\epsilon\in[b,a + b]$, $a, b \geq 0$ and $a + b \leq1$. The player generates a random number $num$ where $num \in[0,1]$. If $num < \epsilon$, the player will explore a random action, else the player will exploit best action from the currently learnt $Q(s,a)$ table. Note that in this function, in order to assess the potential of Q-learning in detail, we introduce $l$ for controlling the decay of $\epsilon$. This parameter determines the value and changing speed of $\epsilon$ in current match count $m$. Instances in our experiments are shown in Fig~\ref{fig:figepsilon}:
\begin{figure}[H]
\centering
\includegraphics[width=0.75\textwidth]{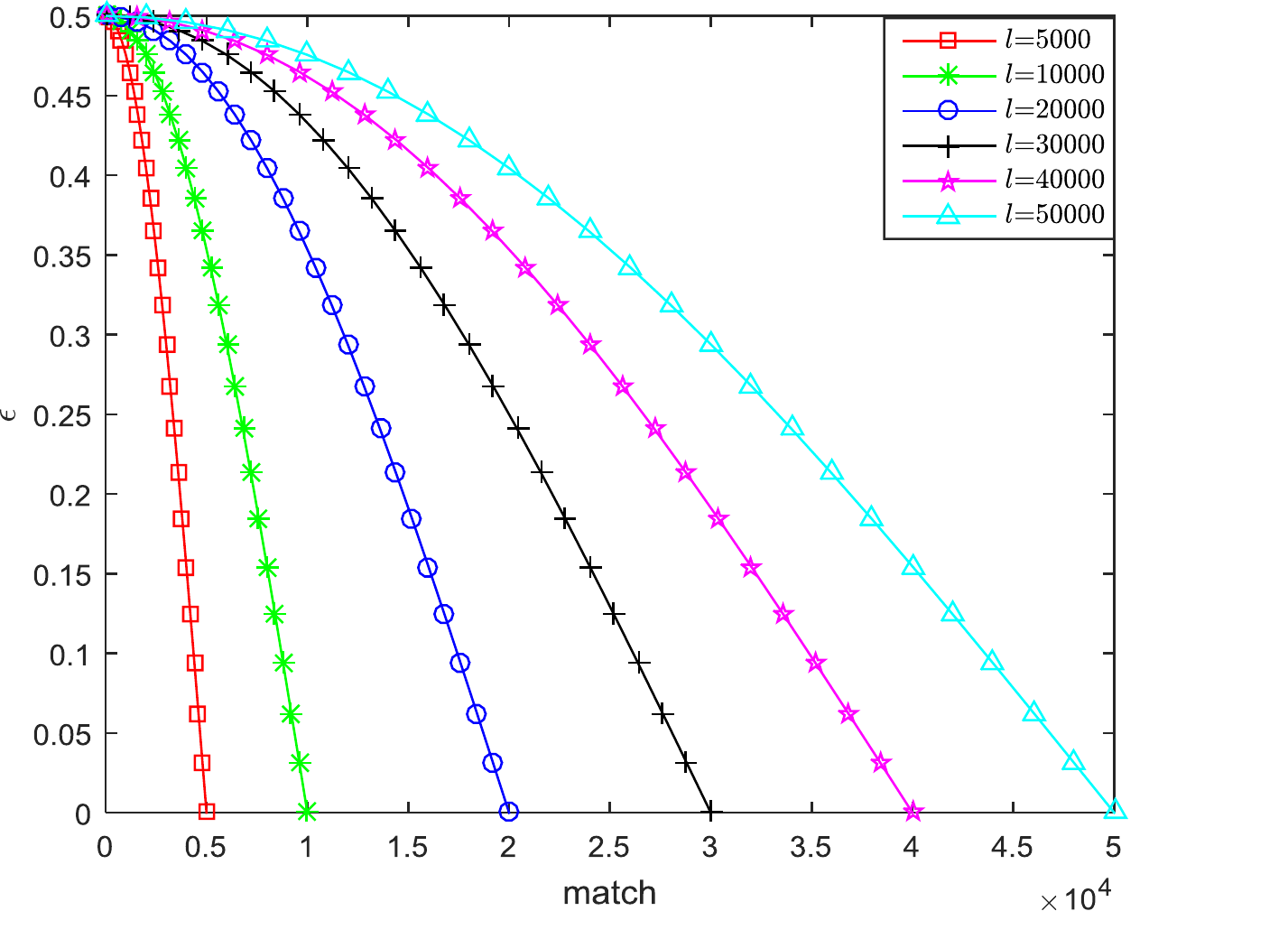}
\caption{Decaying Curves of $\epsilon$ with Different $l$. Every curve decays from 0.5 (learning start, explore $\&$ exploit) to 0 ($m\geq l$, fully exploit).}
\label{fig:figepsilon} 
\end{figure}

\subsection{QM-learning Enhancement}\label{QM-learning}
The main idea of Monte Carlo Search~\cite{Robert2004} is to make some lookahead probes from a non-terminal state to the end of the game by selecting random moves for the players to estimate the value of that state. To apply Monte Carlo in game playing, we use a time-limited version, since in competitive game playing time for each move is an important factor for the player to consider. The time limited MCS in GGP that we use is written as $\emph{\textbf{MonteCarloSearch(time\_limit)}}$.

In Algorithm~\ref{alg:algorithmclassicalQ} (line~\ref{alg:algorithmclassicalQ:random}), we see that a {\em random action\/} is chosen when QPlayer can not find an existing value in the $Q(s,a)$ table. In this case, QPlayer acts like a random player, which will lead to a low win rate and slow learning speed. In order to address this problem, we introduce a variant of Q-learning combined with MCS. MCS performs a time limited lookahead to find better moves. The more time it has, the better the action it finds will be. To achieve this, we use $\emph{\textbf{selected\_action\ =\ MonteCarloSearch(time\_limit)}}$ to replace the line~\ref{alg:algorithmclassicalQ:random}, giving QM-learning. By adding MCS, we effectively add a local version of the last two stages of MCTS to Q-learning: the playout and backup stage~\cite{Browne2012}.

\section{Experiments and Results}
\subsection{Dynamic $\epsilon$ Enhancement}\label{dynamic}
\begin{figure}[H]
\centering
\includegraphics[width=0.75\textwidth]{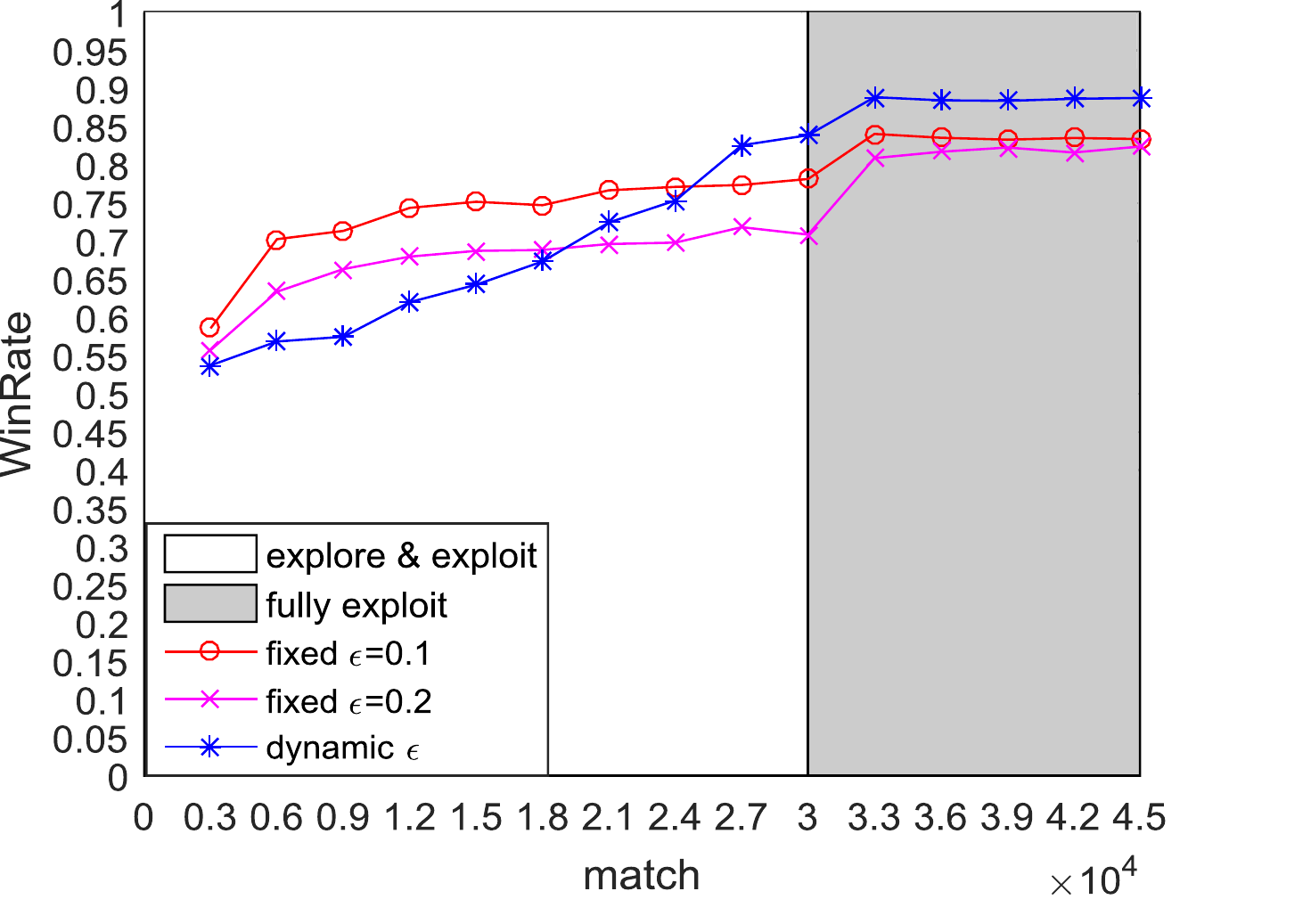}
\caption{Win Rate of the Fixed and Dynamic $\epsilon$ Q-learning Player vs a Random Player Baseline. In the white part, the player uses $\epsilon$-greedy to learn; in the grey part, all players set $\epsilon$=0 (stable performance). The color code of the rest figures are the same}
\label{fig:figfixedepsilon} 
\end{figure}

We create $\epsilon$-greedy Q-learning players~($\alpha=0.1$, $\gamma=0.9$) with fixed $\epsilon$=0.1, 0.2 and with dynamically decreasing $\epsilon \in [0, 0.5]$ to play 30000 matches first ($l$=30000) against a Random player, respectively. During these 30000 matches, the dynamic $\epsilon$  decreases from 0.5 to 0 based on the decay function, see equation~\ref{equation3}. The fixed values for $\epsilon$ are 0.1 and 0.2, respectively. After 30000 matches, fixed $\epsilon$ is also set to 0 to continue the competition. For Tic-Tac-Toe, results in Fig.\ref{fig:figfixedepsilon} show that dynamically decreasing $\epsilon$  performs better. We see that the final win rate of dynamically decreasing $\epsilon$ is 4\%  higher than fixed $\epsilon$=0.1 and 7\% higher than fixed $\epsilon$=0.2. Therefore, in the rest of the experiments, we use dynamic $\epsilon$ for further improvements.

\begin{figure}[H]
\centering
\includegraphics[width=0.75\textwidth]{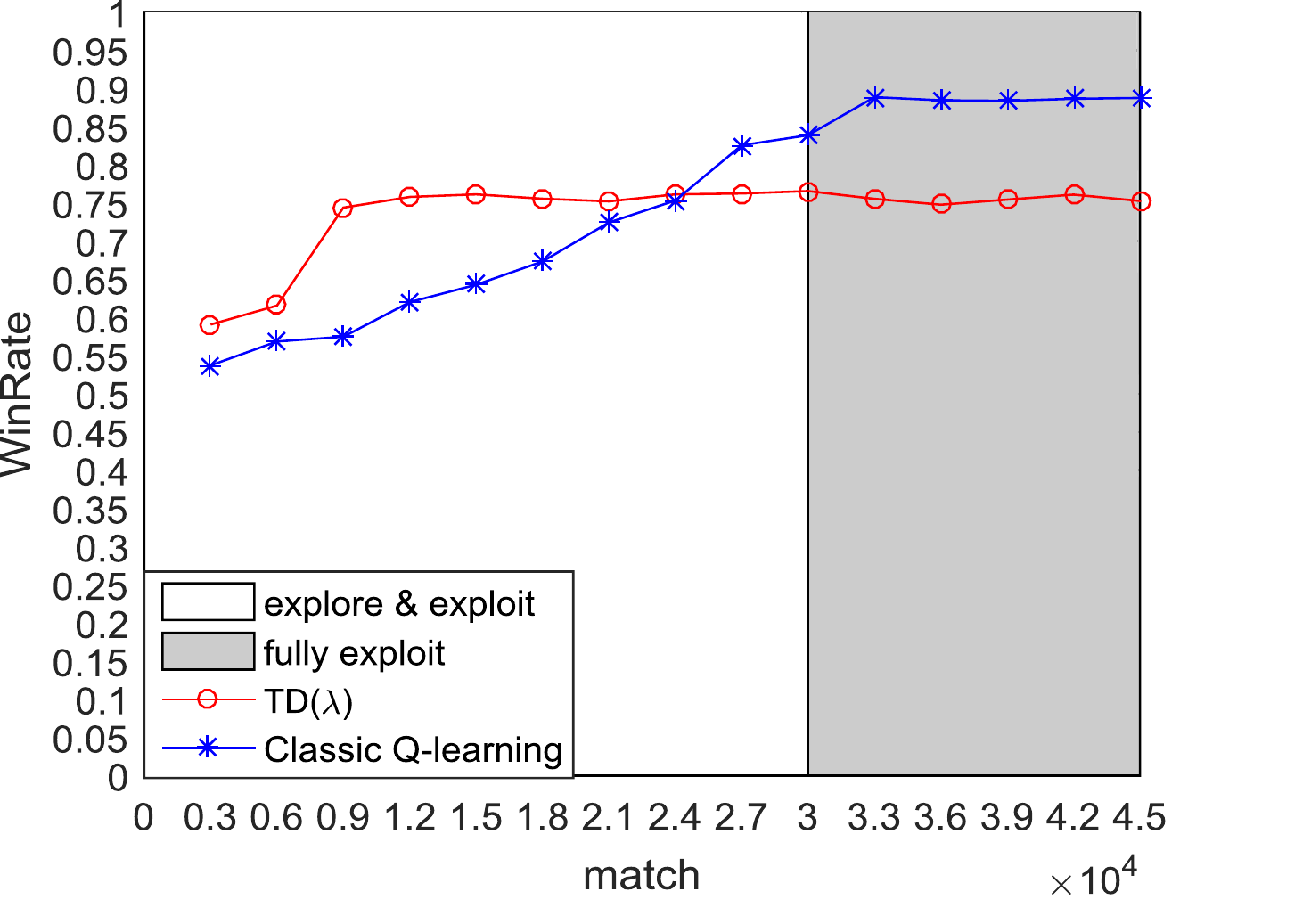}
\caption{Win Rate of Classical Q-learning and [11] Baseline Player vs Random.}
\label{fig:figbaseline} 
\end{figure}

To enable comparison with previous work, we implemented TD($\lambda$), the baseline learner of~\cite{Banerjee2007}($\alpha=0.3$, $\gamma=1.0$, $\lambda=0.7$, $\epsilon=0.01$), and dynamic $\epsilon$ learner($\alpha=0.1$, $\gamma=0.9$, $\epsilon \in [0, 0.5]$, $l$=30000, Algorithm 1). For Tic-Tac-Toe, from Fig.\ref{fig:figbaseline}, we find that although the TD($\lambda$) player converges more quickly initially (win rate stays at about 75.5$\%$ after 9000th match) our dynamic $\epsilon$ player performs better when the value of $\epsilon$ decreases dynamically with the learning process.

\begin{figure}[H]
\centering
\subfigure[3$\times$3 Hex]{\label{fig:subfighex:a} 
\includegraphics[width=0.48\textwidth]{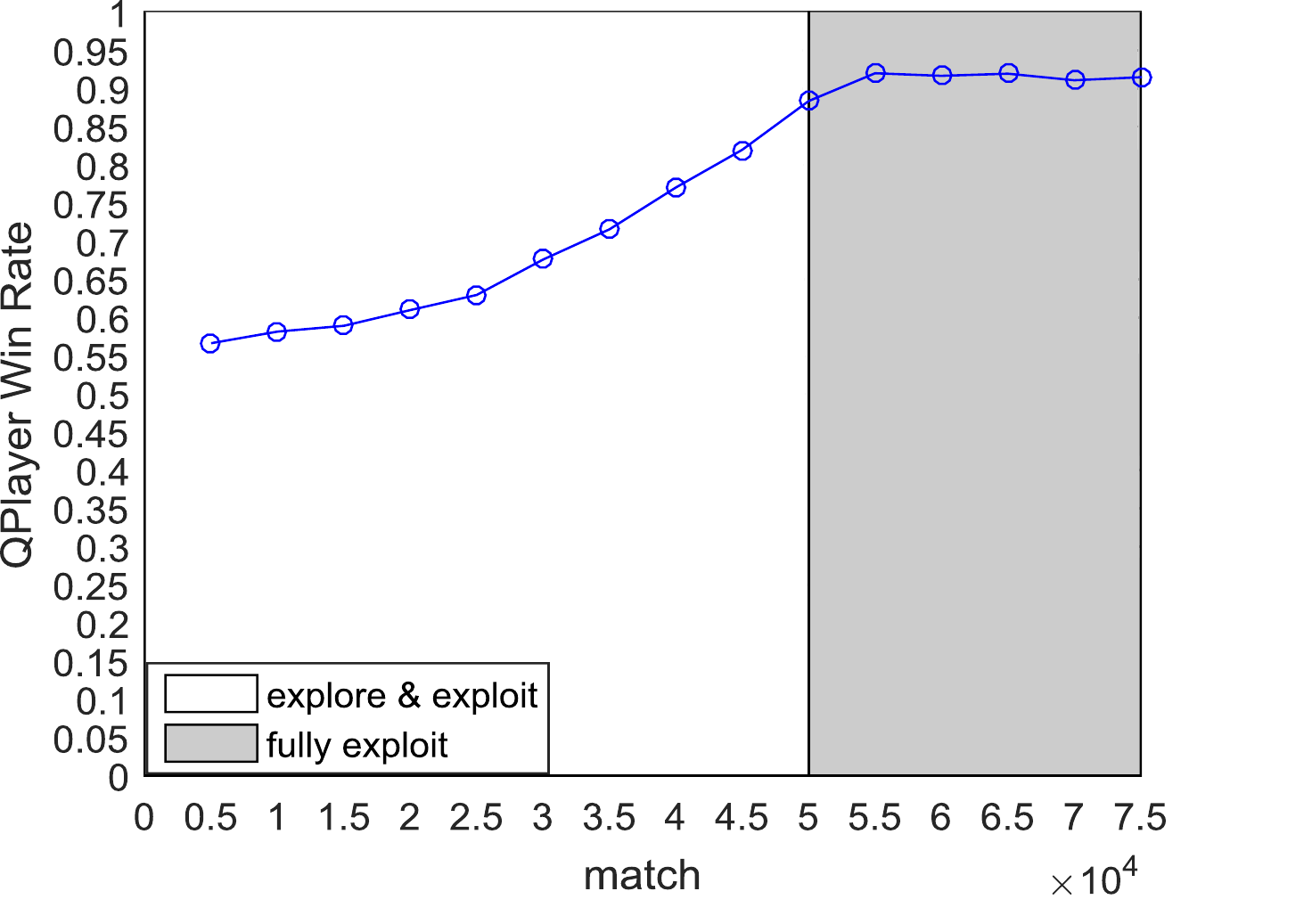}}
\hspace{0.000000000000001\textwidth}
\subfigure[4$\times$4 Connect Four]{\label{fig:subfigconnect4:b} 
\includegraphics[width=0.48\textwidth]{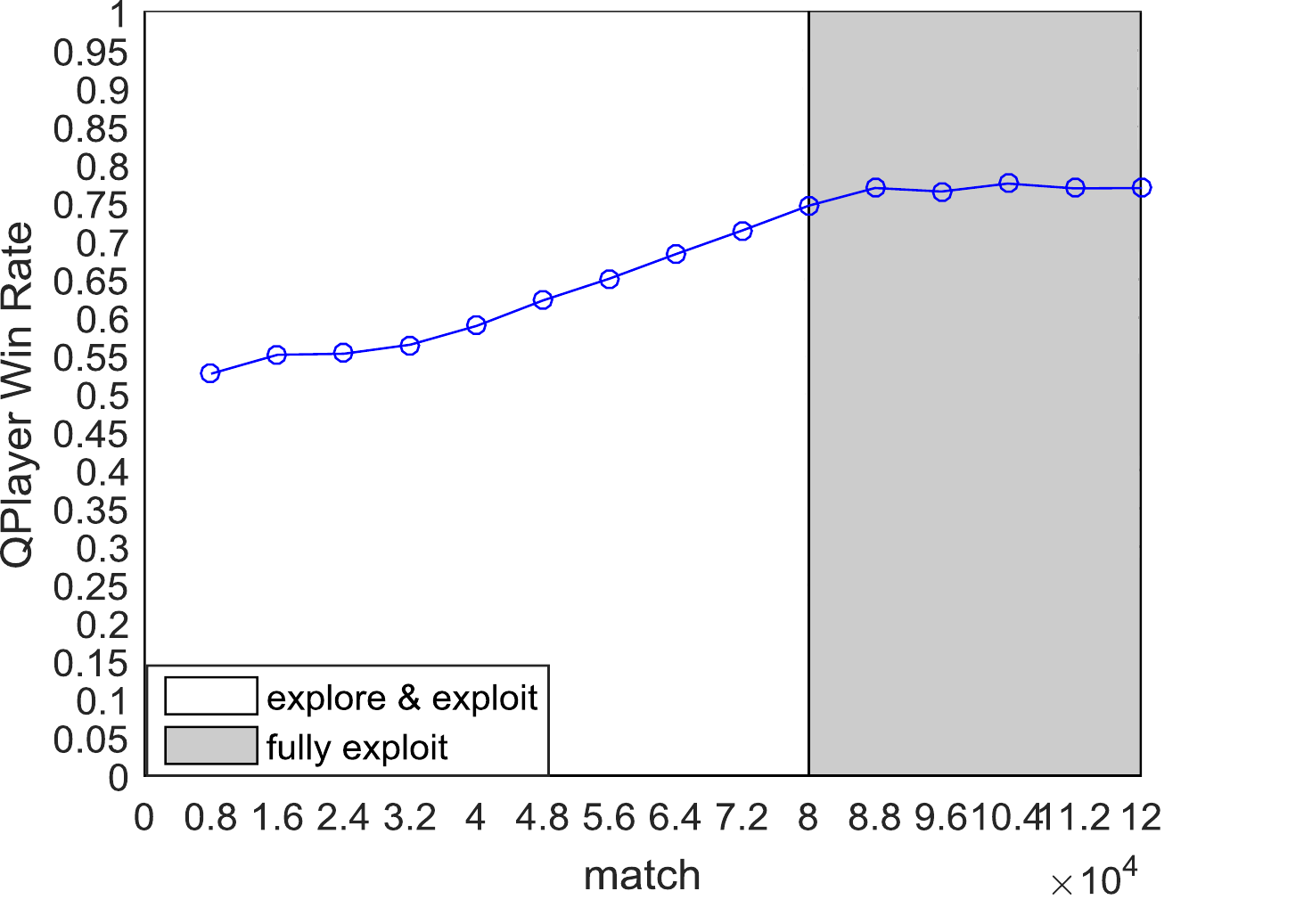}}
\caption{Win Rate of QPlayer vs Random Player in Different Games. For Hex and Connect-Four the win rate of Q-learning also converges}
\label{fig:subfigothergames} 
\end{figure}

Experiments above suggest the following conclusions: that (1) classical Q-learning is applicable to a GGP system, and that (2) a dynamic $\epsilon$ can enhance the performance of fixed $\epsilon$. However, beyond the basic applicability in a single game, we need to show that it can do so (1) {\em efficiently}, and (2) in more than one game. Thus, we further experiment with QPlayer to play Hex ($l$=50000) and Connect Four ($l$=80000) against the Random player. In order to limit excessive learning times, following~\cite{Banerjee2007}, we play Hex on a very small 3$\times$3 board, and play ConnectFour on a 4$\times$4 board. The results of these experiments are given in Fig.\ref{fig:subfigothergames}. We see that QPlayer can also play these other games effectively.



\begin{figure}[H]
\centering
\includegraphics[width=0.75\textwidth]{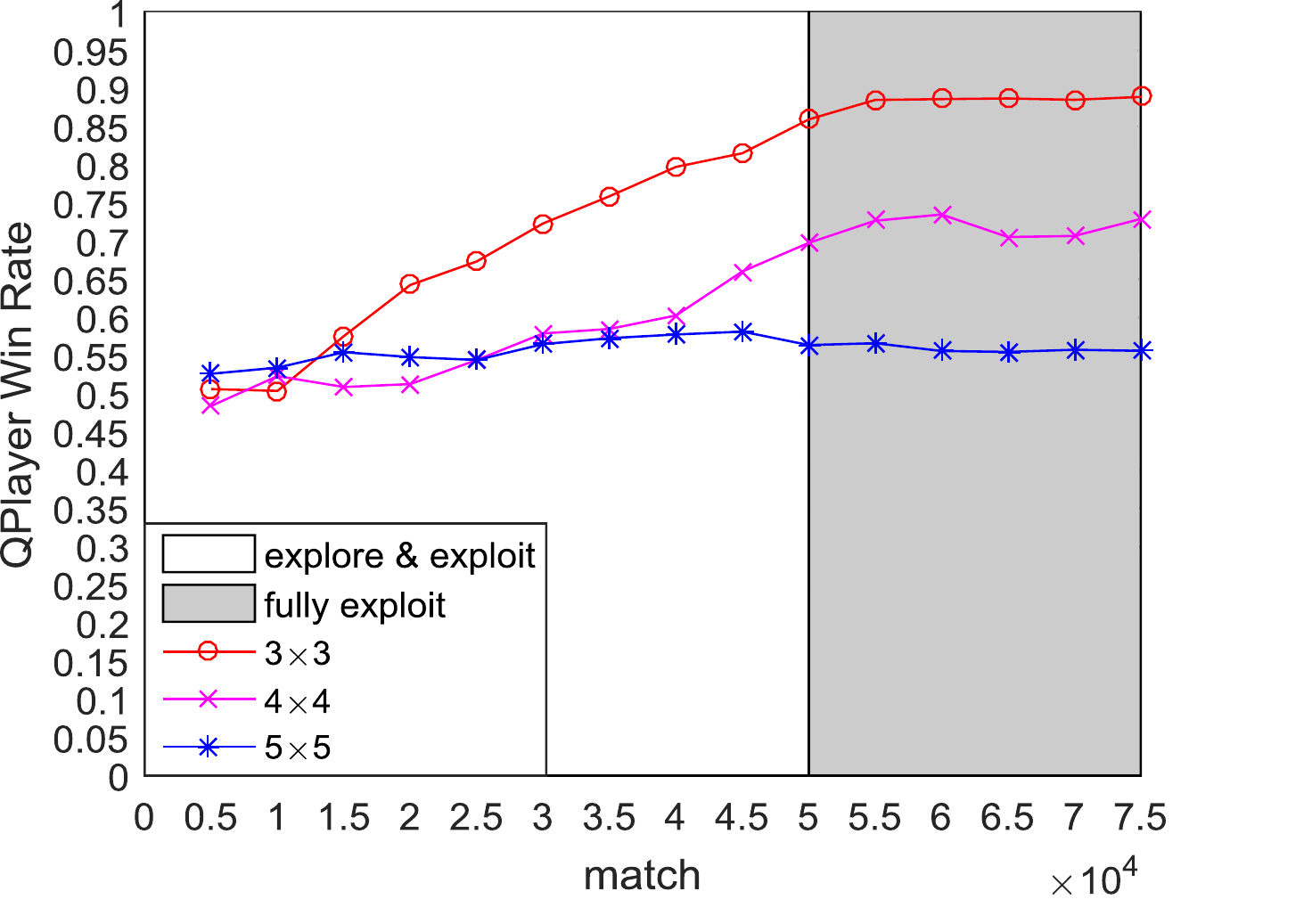}
\caption{Win Rate of QPlayer vs Random in Tic-Tac-Toe on Different Board Size. For larger board sizes convergence slows down
}
\label{fig:figdifferentsize} 
\end{figure}

However, so far, all our games are small. QPlayer should be able to learn to play larger games. The complexity influences how many matches the QPlayer should learn. We will now show results to demonstrate how QPlayer performs while playing more complex games. We make QPlayer play Tic-Tac-Toe~(a line of 3 stones is a win, $l$=50000) in 3$\times$3, 4$\times$4 and 5$\times$5 boards, respectively, and show the results in Fig.\ref{fig:figdifferentsize}.
The results show that with the increase of game board size, QPlayer performs worse.  For larger boards can not achieve convergence. The reason for the lack of convergence is that  QPlayer has not learned enough knowledge. Our experiments also show that for table-based Q-learning in GGP, large game complexity leads to slow convergence, which confirms the well-known  drawback of classical Q-learning.

\subsection{QM-learning Enhancement}
\begin{figure}[H]
\centering
\subfigure[$l$=5000]{\label{fig:subfigQMQ:a} 
\includegraphics[width=0.48\textwidth]{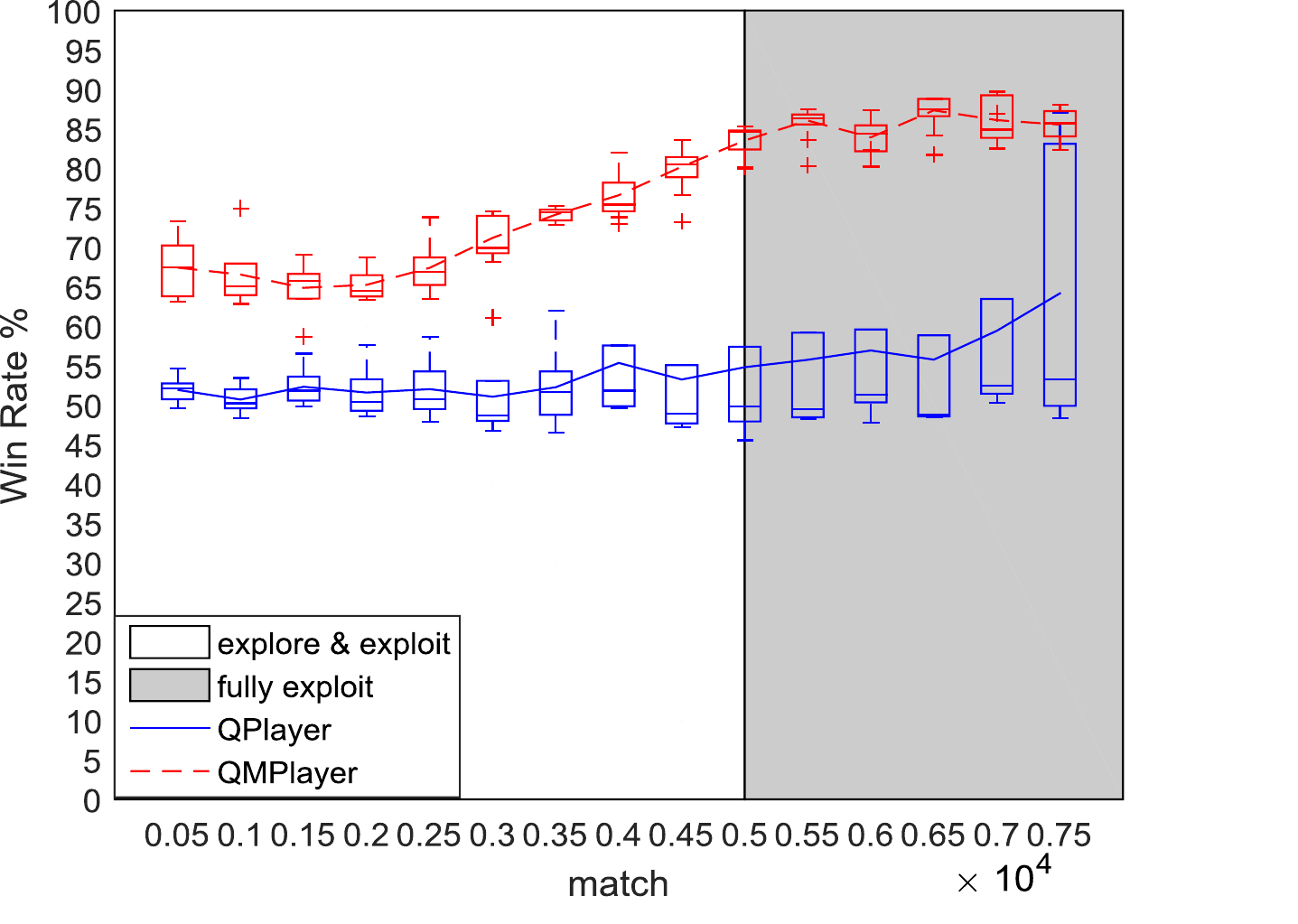}}
\hspace{0.000000000000001\textwidth}
\subfigure[$l$=10000]{\label{fig:subfigQMQ:b} 
\includegraphics[width=0.48\textwidth]{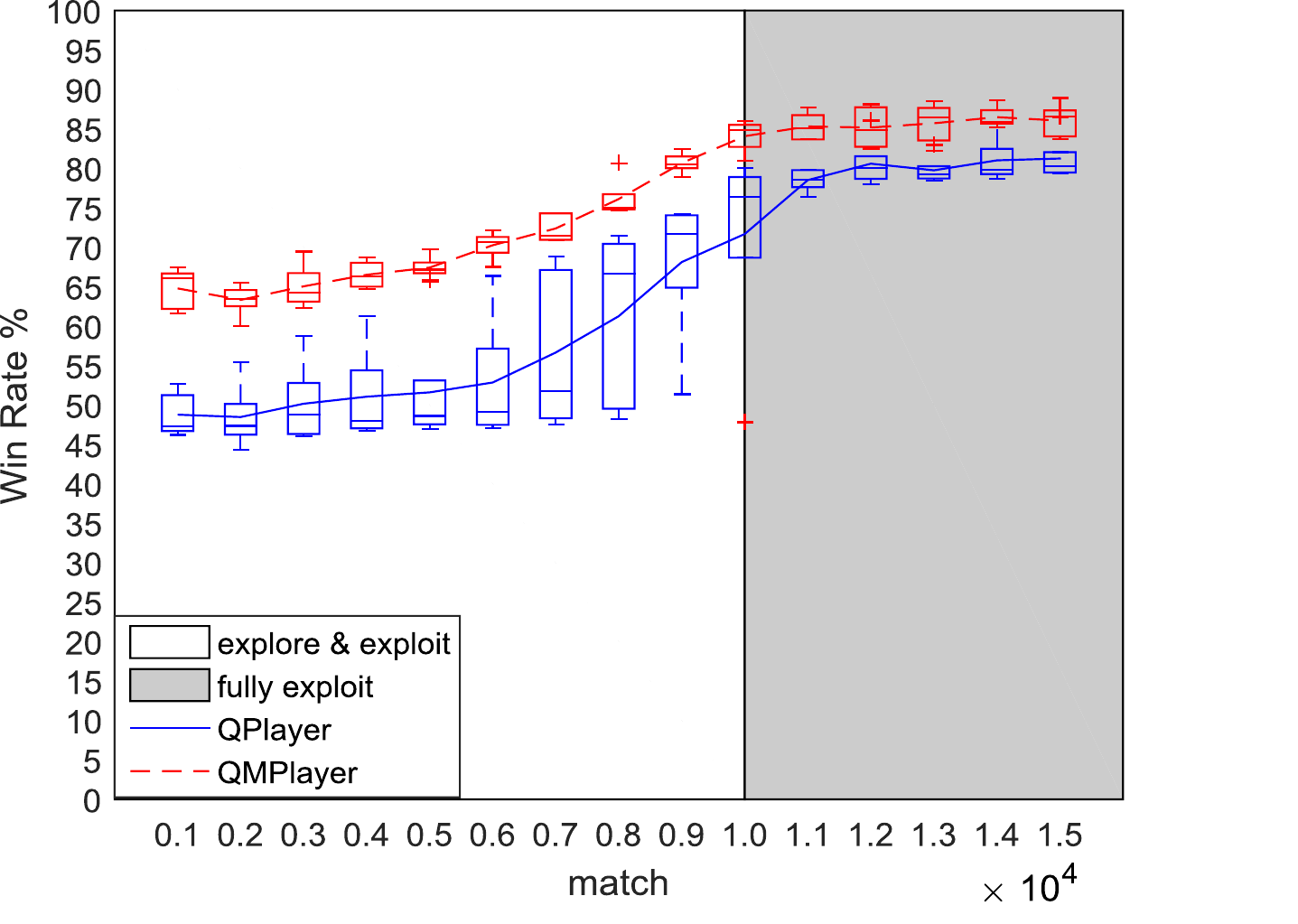}}
\hspace{0.000000000000001\textwidth}
\subfigure[$l$=20000]{\label{fig:subfigQMQ:c} 
\includegraphics[width=0.48\textwidth]{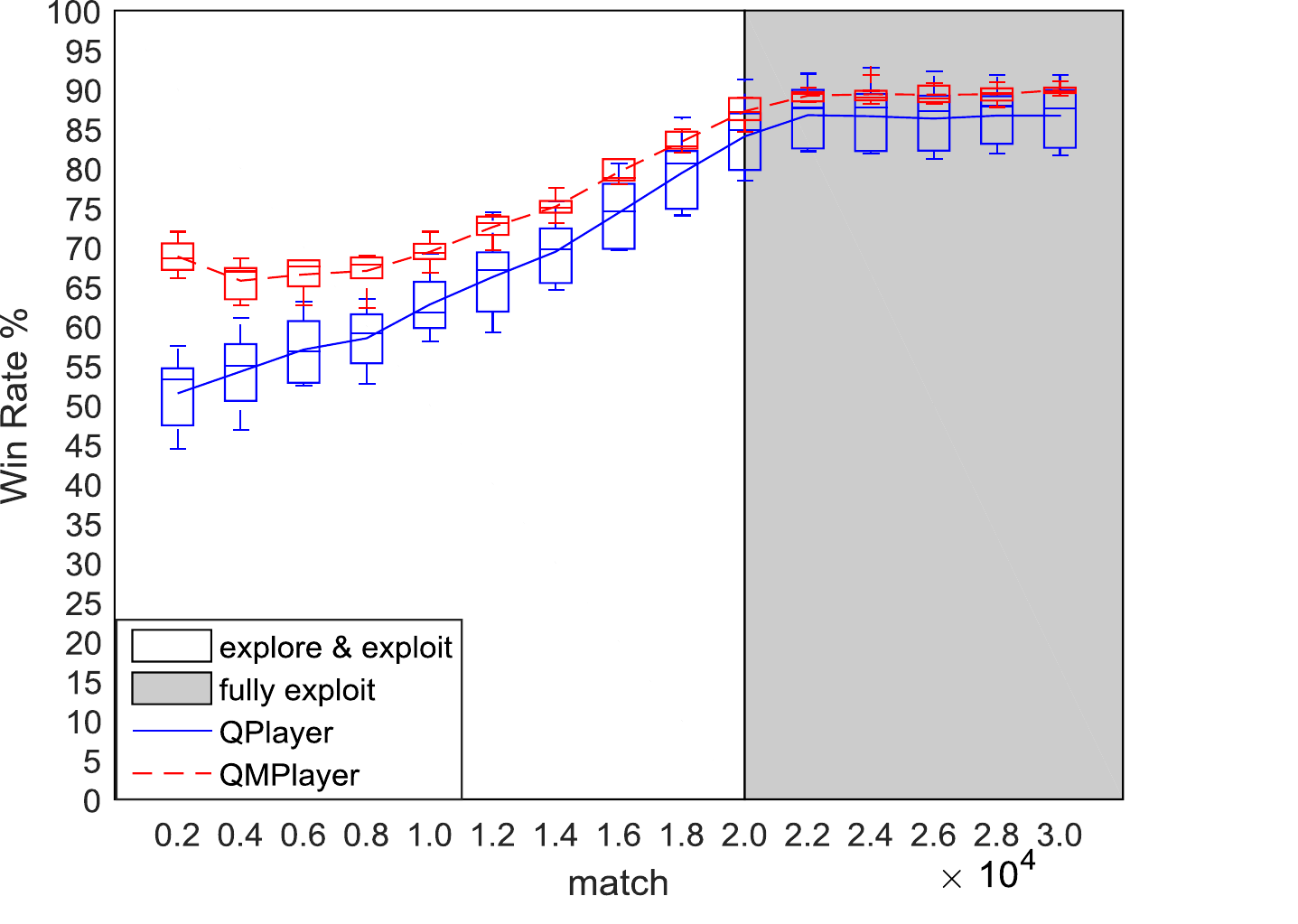}}
\hspace{0.000000000000001\textwidth}
\subfigure[$l$=30000]{\label{fig:subfigQMQ:d} 
\includegraphics[width=0.48\textwidth]{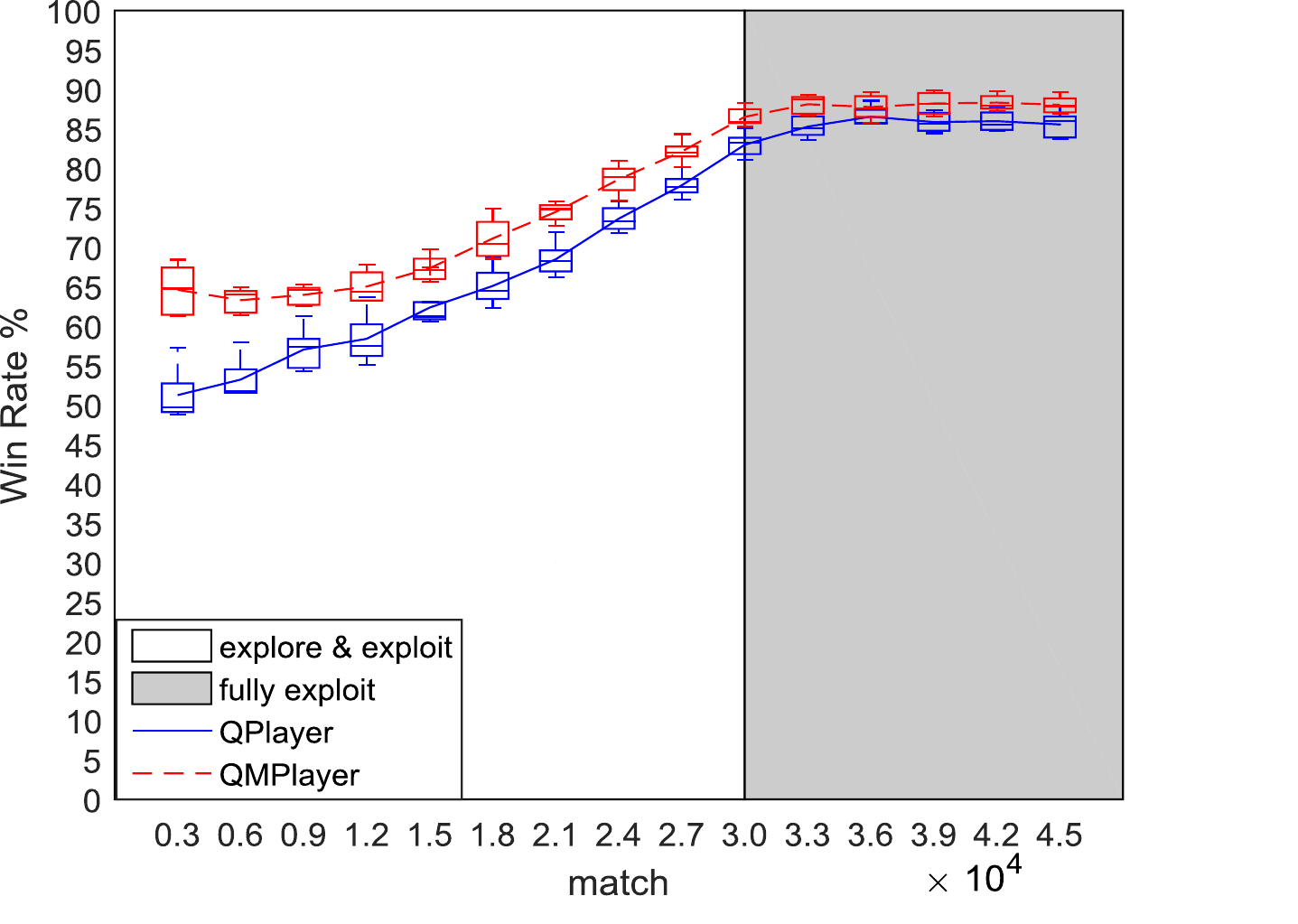}}
\hspace{0.000000000000001\textwidth}
\subfigure[$l$=40000]{\label{fig:subfigQMQ:e} 
\includegraphics[width=0.48\textwidth]{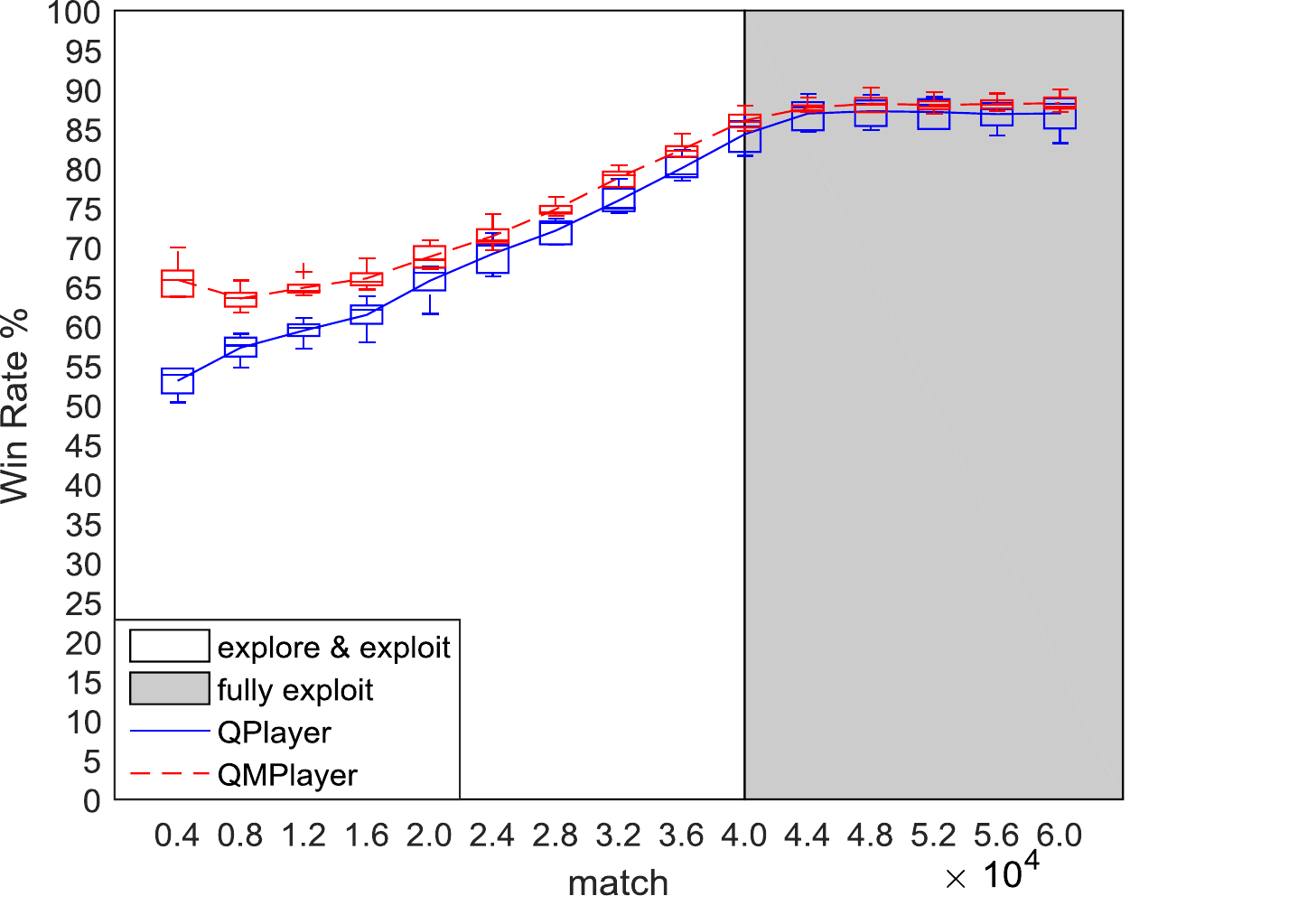}}
\hspace{0.000000000000001\textwidth}
\subfigure[$l$=50000]{\label{fig:subfigQMQ:f} 
\includegraphics[width=0.48\textwidth]{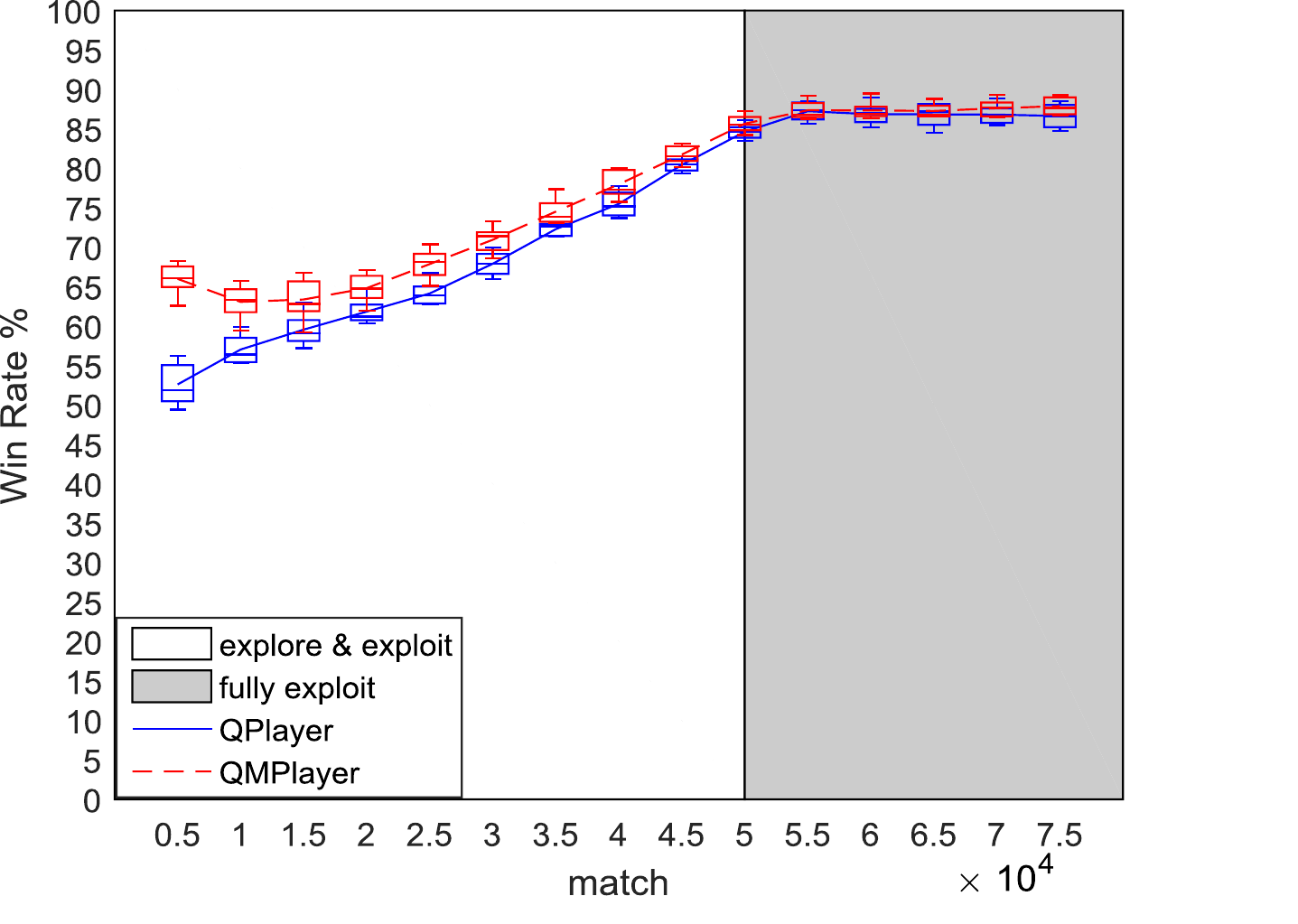}}
\caption{Win Rate of QMPlayer (QPlayer) vs Random in Tic-Tac-Toe for 5 experiments. Small Monte Carlo lookaheads improve the convergence of Q-learning, especially in the early part of learning. QMPlayer always outperforms Qplayer}
\label{fig:subfigQMQ} 
\end{figure}

The second contribution of this paper is QM-learning enhancement, we implement the QPlayer and QMPlayer based on Algorithm~\ref{alg:algorithmclassicalQ} and section~\ref{QM-learning}. For both players, we set parameters to $\alpha=0.1$, $\gamma=0.9$, $\epsilon \in[0, 0.5]$ respectively and we set the $l$=5000, 10000, 20000, 30000, 40000, 50000,  respectively. For QMPlayer, we set $time\_limit=50 ms$. Next we make them play the game with the Random baseline player for $1.5\ \times\ l$ matches for 5 rounds respectively. The comparison between QPlayer and QMPlayer is shown in Fig.\ref{fig:subfigQMQ}.

Fig.\ref{fig:subfigQMQ:a} shows  that QPlayer has the most unstable performance (the largest variance in 5 experiments)  and only wins around 55\% matches after training 5000 matches. Fig.\ref{fig:subfigQMQ:b} illustrates that after training 10000 matches QPlayer wins  about 80\% matches. However, during the exploration period (the white part of the figure) the performance is still very unstable. Fig.\ref{fig:subfigQMQ:c} shows that QPlayer wins about 86\% of the matches while learning 20000 matches still with high variance. Fig.\ref{fig:subfigQMQ:d}, Fig.\ref{fig:subfigQMQ:e}, Fig.\ref{fig:subfigQMQ:f}, show us that after training 30000, 40000, 50000 matches, QPlayer gets a similar win rate, which is nearly 86.5\% with smaller and smaller variance.

In Fig.\ref{fig:subfigQMQ:a}, QMPlayer gets a high win rate~(about 67\%) at the very beginning. Then the win rate decreases to 66\% and 65\%, and then increases from 65\% to around 84\% at the 5000th macth. Finally, the win rate stays at around 85\%. Also in the other sub figures, for QMPlayer, the curves all decrease first and then increase until reaching a stable state. This is because at the very beginning, QMPlayer chooses more actions from MCS. Then as the learning period moves forward, it chooses more actions from Q table.

\begin{figure}[H]
\centering
\includegraphics[width=0.75\textwidth]{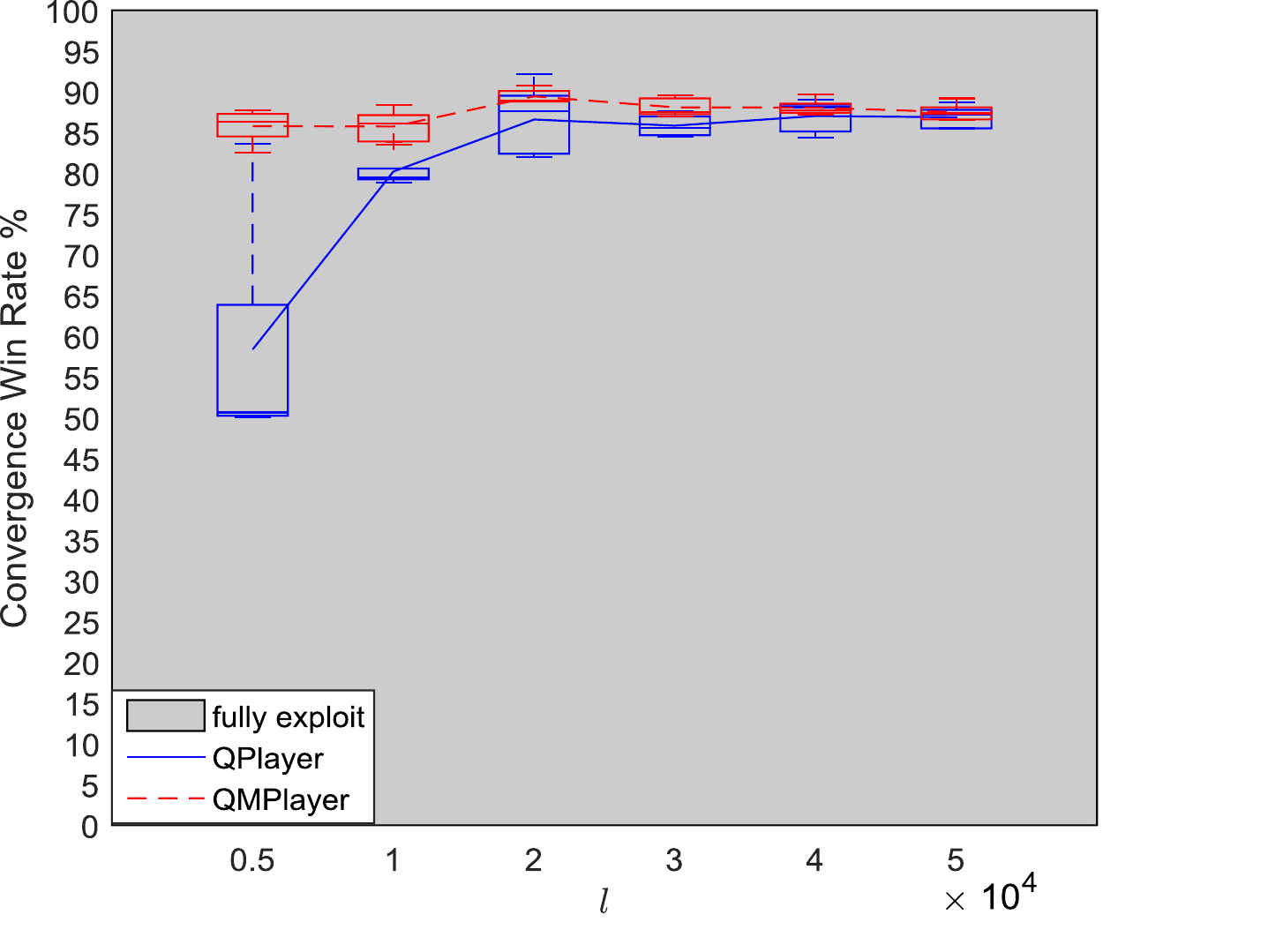}
\caption{Convergence Win Rate of QMPlayer (QPlayer) vs Random in Tic-Tac-Toe}
\label{fig:figsummarize} 
\end{figure}

Overall, as the $l$ increases, the win rate of QPlayer becomes higher until leveling off around 86.5\%. The variance becomes smaller and smaller, which proves that Q-learning can achieve convergence in GGP games and that a proper $\epsilon$ decaying speed makes sense for classical Q-learning. Note that in every sub figure, QMPlayer can always achieve a higher win rate than QPlayer, not only at the beginning but also at the end of the learning period. Overall, QMPlayer achieves a better performance than QPlayer with the higher convergence win rate (at least 87.5\% after training 50000 matches). To compare the convergence speeds of QPlayer and QMPlayer, we summarize the convergence win rates of different $l$ according to Fig.\ref{fig:subfigQMQ} in Fig.\ref{fig:figsummarize}.

These results show that combining online MCS with classical Q-learning for GGP can improve the win rate both at the beginning and at the end of the offline learning period. The main reason is that QM-learning allows the $Q(s,a)$ table to be filled quickly with good actions from MCS, achieving a quick and direct learning rate. It is worth to note that,  QMPlayer will spend slightly more time (at most is $search~time~limit\times$ {\em number~of~(state-action)~pairs}) in training than QPlayer. It will be time consuming for MCS to compute a large game, and this is also the essential drawback of table-based Q-learning, so currently QM-learning is also only applicable for small games.

\section{Conclusion}
This paper examines the applicability of Q-learning, a canonical reinforcement learning algorithm, to create general players for GGP programs. Firstly, we show how good canonical implementations of Q-learning perform on GGP games. The GGP system allows us to easily use three real games for our experiments: Tic-Tac-Toe, Connect Four, and Hex. We find that (1) Q-learning is indeed general enough to achieve convergence in GGP games. However, we also find that convergence is slow. In accordance with Banerjee~\cite{Banerjee2007}, who used a static value for $\epsilon$, we find that (2) a value for $\epsilon$ that changes with the learning phases gives better performance (start with more exploration, become more greedy later on). The table-based implementation of Q-learning facilitates theoretical analysis, and comparison against some baselines \cite{Banerjee2007}. However, it is only suitable for small games. A neural network implementation facilitates the study of larger games, and allows meaningful comparison to DQN variants \cite{Mnih2015}.

Still using our table-based implementation, we then enhance Q-learning with an MCS based lookahead. We find that, especially at the start of the learning, this speeds up convergence considerably. Our Q-learning is table-based, limiting it to small games. Even with the MCS enhancement, convergence of QM-learning does not yet allow its direct use in larger games. The QPlayer needs to learn a large number of matches to get good performance in playing larger games. The results with the improved Monte Carlo algorithm show a real improvement of the player's win rate, and learn the most probable strategies to get high rewards faster than learning completely from scratch. This enhancement shows how online search can be used to improve the performance of offline learning in GGP. On this basis, we can assess different offline learning algorithms (or follow Gelly~\cite{Gelly2007} to combine it with neural networks for larger games in GGP).

Our use of Monte Carlo in QM-learning is different from the AlphaGo architecture, where MCTS is wrapped around Q-learning (DQN) \cite{Mnih2015}. In our approach, we insert Monte Carlo {\em within} the Q-learning loop.
Future work should show if our QM-learning results transfer to AlphaGo-like uses of DQN inside MCTS, if QM-learning can  achieve faster convergence, reducing the high computational demands of AlphaGo \cite{Silver2017a}. Additionally, we plan to study  nested MCS in Q-learning \cite{Cazenave2016}. Implementing Neural Network based players also allows the study of more complex GGP games.

\subsubsection*{Acknowledgments.} Hui Wang acknowledges financial support from the China Scholarship Council (CSC), CSC No.201706990015.

\end{document}